\documentclass[10pt,twocolumn,letterpaper]{article}

\usepackage{cvpr}              

\usepackage{graphicx}
\usepackage{amsmath}
\usepackage{amssymb}
\usepackage{booktabs}
\usepackage{bbding}
\usepackage{times}
\usepackage{epsfig}
\usepackage{multirow}
\usepackage{threeparttable}  
\usepackage{color}
\usepackage{array}
\usepackage{makecell}
\usepackage{stfloats}
\usepackage[table]{xcolor}
\usepackage[misc]{ifsym}
\usepackage[accsupp]{axessibility}

\newcommand\blfootnote[1]{%
  \begingroup
  \renewcommand\thefootnote{}\footnote{#1}%
  \addtocounter{footnote}{-1}%
  \endgroup
}
\usepackage[pagebackref,breaklinks,colorlinks]{hyperref}

\def\cvprPaperID{8613} 
\def\confName{CVPR}
\def\confYear{2022}

\begin{document}

\title{MixFormer: End-to-End Tracking with Iterative Mixed Attention}

\author{Yutao Cui \quad \quad Cheng Jiang \quad \quad Limin Wang \textsuperscript{\Letter} \quad \quad Gangshan Wu\\
State Key Laboratory for Novel Software Technology,
Nanjing University,
China \\
{\tt\small \{cuiyutao,mg1933027\}@smail.nju.edu.cn} \quad
{\tt\small \{lmwang,gswu\}@nju.edu.cn}
}

\maketitle
\blfootnote{\Letter~: Corresponding author.}
\begin{abstract}
Tracking often uses a multi-stage pipeline of feature extraction, target information integration, and bounding box estimation. To simplify this pipeline and unify the process of feature extraction and target information integration, we present a compact tracking framework, termed as {\em MixFormer}, built upon transformers. Our core design is to utilize the flexibility of attention operations, and propose a Mixed Attention Module (MAM) for simultaneous feature extraction and target information integration. This synchronous modeling scheme allows to extract target-specific discriminative features and perform extensive communication between target and search area. Based on MAM, we build our MixFormer tracking framework simply by stacking multiple MAMs with progressive patch embedding and placing a localization head on top. In addition, to handle multiple target templates during online tracking, we devise an asymmetric attention scheme in MAM to reduce computational cost, and propose an effective score prediction module to select high-quality templates. 
Our MixFormer sets a new state-of-the-art performance on five tracking benchmarks, including LaSOT, TrackingNet, VOT2020, GOT-10k, and UAV123. 
In particular, our MixFormer-L achieves NP score of 79.9\% on LaSOT, 88.9\% on TrackingNet and EAO of 0.555 on VOT2020.
We also perform in-depth ablation studies to demonstrate the effectiveness of simultaneous feature extraction and information integration. Code and trained models are publicly available at \href{https://github.com/MCG-NJU/MixFormer}{https://github.com/MCG-NJU/MixFormer}.

\end{abstract}

\section{Introduction}





Visual object tracking~\cite{updt,mdnet,dcf_,CRPN,bacf,updt,staple,vital} has been a fundamental task in computer vision area for decades, aiming to estimate the state of an arbitrary target in video sequences given its initial status. It has been successfully deployed in various applications such as human computer interaction~\cite{introduction1} and visual surveillance~\cite{introduction2}. However, how to design a simple yet effective end-to-end tracker is still challenging in real-world scenarios. The main challenges are from aspects of scale variations, object deformations, occlusion, and confusion from similar objects.

\begin{figure}[t]
\centering
\includegraphics[width=\linewidth]{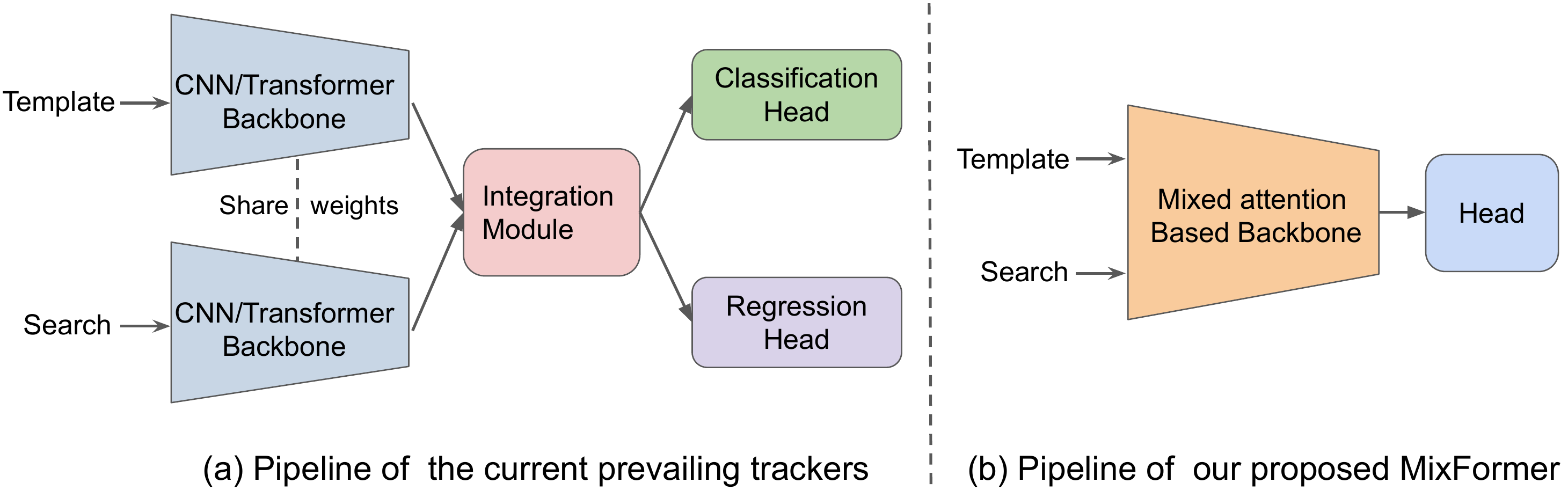}
\caption{{\bf Comparison of tracking pipeline}. (a) The dominant tracking framework contains three components: a convolutional or transformer backbone, a carefully-designed integration module, and task-specific heads. (b) Our MixFormer is more compact and composed of two components: a target-search mixed attention based backbone and a simple localization head. 
}
\vspace{-3mm}
\label{fig:motivation}
\end{figure}

Current prevailing trackers typically have a multi-stage pipeline as shown in Fig.~\ref{fig:motivation}. It contains several components to accomplish the tracking task: (1) a backbone to extract generic features of tracking target and search area, (2) an integration module to allow information communication between tracking target and search area for subsequent target-aware localization, (3) task-specific heads to precisely localize the target and estimate its bounding box. Integration module is the key of tracking algorithms as it is responsible for incorporating the target information to bridge the steps of generic feature extraction and target-aware localization. 
Traditional integration methods include correlation-based operations (e.g. SiamFC~\cite{siamfc}, SiamRPN~\cite{siamrpn}, CRPN~\cite{CRPN}, SiamFC++~\cite{siamfc++}, SiamBAN~\cite{siamban}, OCEAN~\cite{ocean}) and online learning algorithms (e.g., DCF~\cite{dcf_}, KCF~\cite{kcf}, CSR-DCF~\cite{csr_dcf}, ATOM~\cite{atom}, DiMP~\cite{dimp}, FCOT~\cite{fcot}). Recently, thanks to its global and dynamic modeling capacity, transformers~\cite{transformer} are introduced to perform attention based integration and yields good tracking performance (e.g., TransT~\cite{tt}, TMT~\cite{tmt}, STMTrack~\cite{stmtrack}, TREG~\cite{treg}, STARK~\cite{stark}, DTT~\cite{dtt}). 
However, these transformer based trackers still depend on the CNN for generic feature extraction, and only apply attention operations in the latter high-level and abstract representation space. We analyze that these CNN representations are limited as they are typically pre-trained for generic object recognition and might neglect finer structure information for tracking. In addition, these CNN representations employ local convolutional kernels and lack global modeling power. Therefore, CNN representation is still their bottleneck, which prevents them from fully unleashing power of self-attention for the whole tracking pipeline.

To overcome the above issue, we present a new perspective on tracking framework design that generic feature extraction and target information integration should be coupled together within a unified framework. This coupled processing paradigm shares several key advantages. First, it will enable our feature extraction to be more specific to the corresponding tracking target and capture more target-specific discriminative features. Second, it also allows the target information to be more extensively integrated into search area, and thereby to better capture their correlation. In addition, this will result in a more compact and neat tracking pipeline only with a single backbone and tracking head, without an explicit integration module.

Based on the above analysis, in this paper, we introduce the \emph{MixFormer}, a simple tracking framework designed for unifying the feature extraction and target integration solely with a transformer-based architecture. Attention module is a very flexible architectural building block with dynamic and global modeling capacity, which makes few assumption about the data structure and could be generally applied for general relation modeling. Our core idea is to utilize this flexibility of attention operation, and present a {\em mixed attention module} (MAM) that performs both of feature extraction and mutual interaction of target template and search area at the same time. In particular, in our MAM, we devise a hybrid interaction scheme with both self-attention and cross-attention operations on the tokens from target template and search area. The self-attention is responsible to extract their own features of target or search area, while the cross-attention allows for the communications between them to mix the target and search area information. To reduce computational cost of MAM and thereby allow for multiple templates to handle object deformation, we further present a customized {\em asymmetric} attention scheme by pruning the unnecessary target-to-search area cross-attention.

Following the successful transformer architecture in image recognition, we build our MixFormer backbone by stacking the layers of Patch Embedding and MAM, and finally place a simple localization head to yield our whole tracking framework. 
As a common practice in dealing with object deformation during tracking procedure, we also propose a score based target template update mechanism and our MixFormer could be easily adapted for multiple target template inputs. Extensive experiments on several benchmarks demonstrate that \textit{MixFormer} sets a new state-of-the-art performance, with a real-time running speed of 25 FPS on a GTX 1080Ti GPU.
Especially, \textit{MixFormer-L} surpasses STARK~\cite{stark} by 5.0\% (EAO score) on VOT2020, 2.9\% (NP score) on LaSOT and 2.0\% (NP score) on TrackingNet.

The main contributions are summarized as follows:
\begin{itemize}
\item We propose a compact end-to-end tracking framework, termed as \emph{MixFormer}, based on iterative Mixed Attention Modules (MAM). It allows for extracting target-specific discriminative features and extensive communication between target and search simultaneously.
\item For online template update, we devise a customized asymmetric attention in MAM for high efficiency, and propose an effective score prediction module to select high-quality templates, leading to an efficient and effective online transformer-based tracker. 
\item The proposed \textit{MixFormer} sets a new state-of-the-art performance on five challenging benchmarks, including VOT2020~\cite{vot2020}, LaSOT~\cite{lasot}, TrackingNet~\cite{trackingnet}, GOT-10k~\cite{got10k}, and UAV123~\cite{uav123}.

\end{itemize}

\section{Related Work}
\paragraph{Tracking Paradigm.}
Current prevailing tracking methods can be summarized as a three-parts architectures, containing (\romannumeral1) a backbone to extract generic features, (\romannumeral2) an integration module to fuse the target and search region information, (\romannumeral3) heads to produce the target states. Generally, most trackers~\cite{siamrpnPlus, atom,dimp,transt,siamrcnn} used ResNet as the backbone. For the most important integration module, researchers explored various methods. Siamese-based trackers~\cite{siamfc,siamrpn,siamcar,dasiamrpn} combined a correlation operation with the Siamese network, modeling the global dependencies between the target and search. Some online trackers~\cite{eco,kcf,bacf,strcf,dcf_,atom,dimp,fcot} learned an target-dependent model for discriminative tracking. Furthermore, some recent trackers~\cite{transt,stark,treg,stmtrack,tmt} introduced a transformer-based integration module to capture more complicated dependencies and achieved impressive performance. Instead, we propose a fully end-to-end transformer tracker, solely containing a MAM based backbone and a simple head, leading to a more accurate tracker with neat and compact architecture.

\paragraph{Vision Transformer.} 
The Vision Transformer (ViT)~\cite{vit} first presented a pure vision transformer architecture, obtaining an impressive performance on image classification.
Some works~\cite{t2t,localvit,pvt} introduced design changes to better model local context in vision Transformers. For example, PVT~\cite{pvt} incorporated a multi-stage design (without convolutions) for Transformer similar to multi-scales in CNNs.
CVT~\cite{cvt} combined CNNs and Transformers to model both local and global dependencies for image classification in an efficient way. Our MixFomer uses the pre-trained CVT models, but there are some fundamental differences. (\romannumeral1) The proposed MAM performs dual attentions for both feature extraction and information integration, while CVT uses self attention to solely extract features. (\romannumeral2) The learning tasks are different, and the corresponding input and the head are different. We use multiple templates together with the search region as input and employ a corner-based or query-based localization head for bounding box generation, while CVT is designed for image classification. (\romannumeral3) We further introduce an asymmetric mixed attention and a score prediction module for the specific task of online tracking. 

Attention machenism has been also explored in object tracking recently. CGCAD~\cite{CGACD} and SiamAttn~\cite{siamattn} introduced a correlation-guided attention and self-attention to perform discriminative tracking. TransT~\cite{transt} designed a transformer-based fusion network for target-search information incorporation. These methods still relied on post-processing for box generation. Inspired by DETR~\cite{detr}, STARK~\cite{stark} further proposed an end-to-end transformer-based tracker. However, it still followed the paradigm of \emph{Backbone-Integration-Head}, with separate feature extraction and information integration modules.
Meanwhile, TREG~\cite{treg} proposed a target-aware transformer for regression branch and can generate accurate prediction in VOT2021~\cite{vot2021}. Inspired by TREG, we formulate mixed attention mechanism by using both self attention and cross attention.
In this way, our MixFormer unifies the two processes of feature extraction and information integration with an iterative MAM based backbone, leading to a more compact, neat and effective end-to-end tracker.

\begin{figure*}[pt]
\centering
\includegraphics[width=0.8\linewidth]{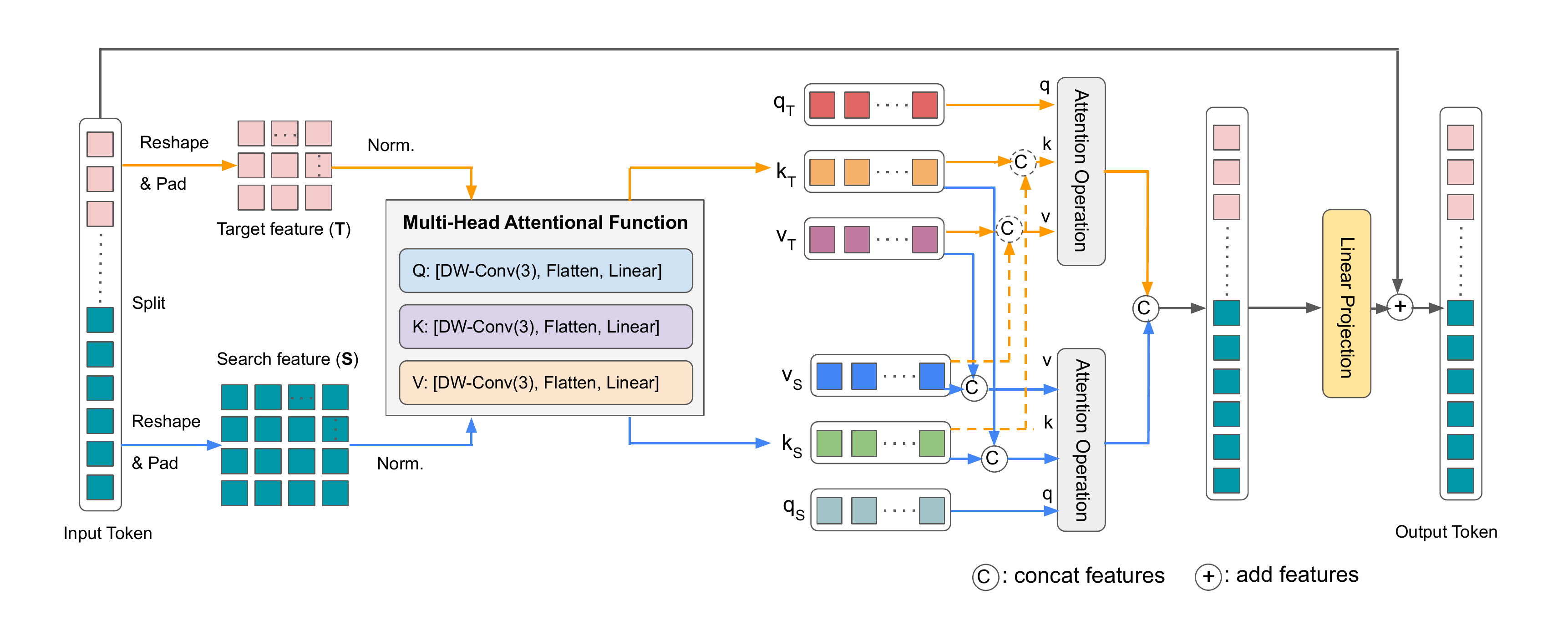}
\vspace{-3mm}
\caption{{\bf Mixed Attention Module (MAM)} is a flexible attention operation that unifies the process of feature extraction and information integration for target template and search area. This mixed attention has dual attention operations where self-attention is performed to extract features from itself while cross-attention is conducted to communicate between target and search. This MAM could be easily implemented with a concatenated token sequence. To further improve efficiency, we propose an asymmetric MAM by pruning the target-to-search cross attention (denoted by dashed lines).
}
\vspace{-4mm}
\label{fig_mam}
\end{figure*}

\section{Method}
In this section, we present our end-to-end tracking framework, termed as \emph{MixFormer}, based on iterative mixed attention modules (MAM). First, we introduce our proposed MAM to unify the process of feature extraction and target information incorporation. This simultaneous processing scheme will enable our feature extraction to be more specific to the corresponding tracking target. In addition, it also allows the target information integration to be performed more extensively and thus to better capture the correlation between target and search area. Then, we present the whole tracking framework of MixFormer, which only includes a MAM-based backbone and localization head. Finally, we describe the training and inference of MixFormer by devising a confidence score based target template update mechanism to handle object deformation in tracking procedure.

\subsection{Mixed Attention Module (MAM)}
\label{sec:MAM}

\begin{figure*}[t]
\centering
\includegraphics[width=0.85\linewidth]{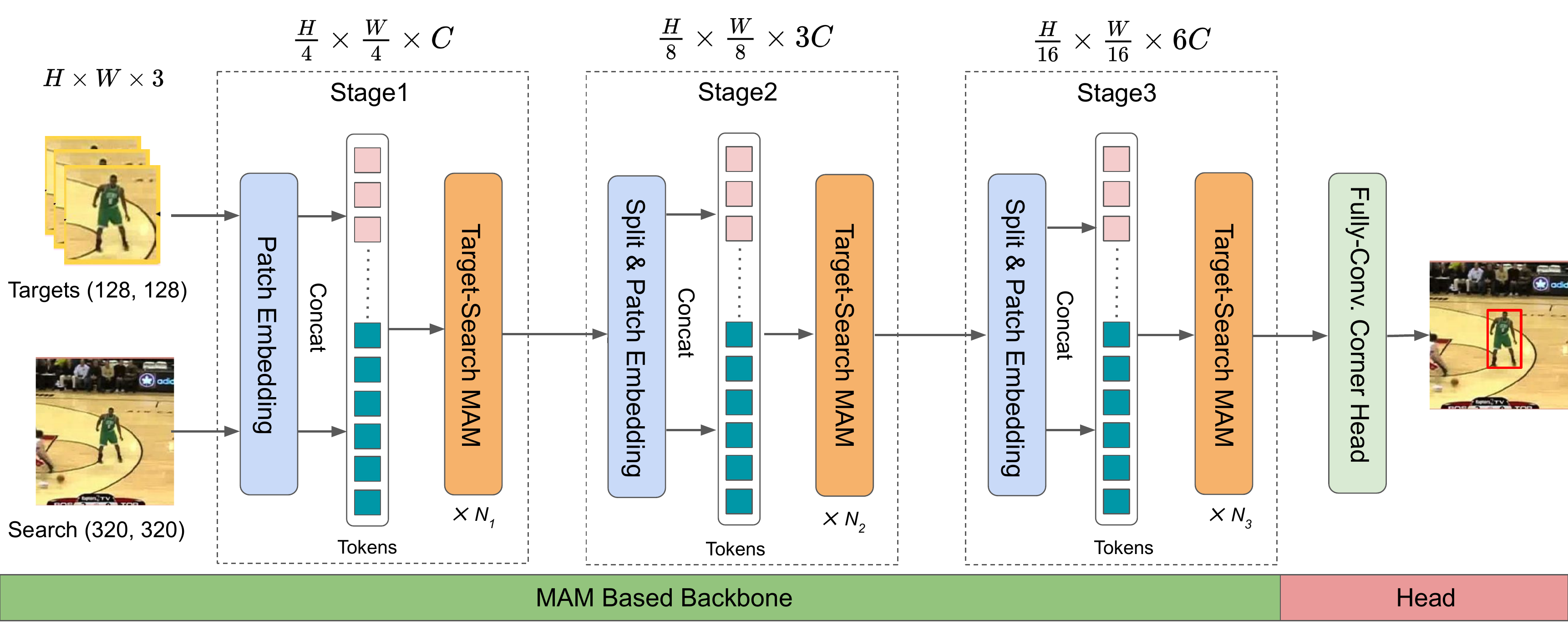}
\vspace{-2mm}
\caption{{\bf MixFormer} presents a compact end-to-end framework for tracking without explicitly decoupling steps of feature extraction and target information integration. It is only composed of a single MAM backbone and a localization head.}
\label{fig_arch}
\vspace{-6mm}
\end{figure*}

Mixed attention module (MAM) is the core design to pursue a neat and compact end-to-end tracker. The input to our MAM is the {\em target template} and {\em search area}. It aims to simultaneously extract their own long-range features and fuse the interaction information between them.
In contrast to the original Multi Head Attention~\cite{transformer}, MAM performs dual attention operations on two separate tokens sequences of target template and search area. It carries out self-attention on tokens in each sequence themselves to capture the target or search specific information. Meanwhile, it conducts cross-attention between tokens from two sequences to allow communication between target template and search area. As shown in Fig.~\ref{fig_mam}, this mixed attention mechanism could be implemented efficiently via a concatenated token sequence.

Formally, given a concatenated tokens of multiple targets and search, we first split it into two parts and reshape them to 2D feature maps. In order to achieve additional modeling of local spatial context, a separable depth-wise convolutional projection layer is performed on each feature map (i.e., \emph{query}, \emph{key} and \emph{value}). It also provides efficiency benefits by allowing the down-sampling in \emph{key} and \emph{value} matrices. Then each feature map of target and search is flattened and processed by a linear projection to produce queries, keys and values of the attention operation. We use $q_t$, $k_t$ and $v_t$ to represent target, $q_s$, $k_s$ and $v_s$ to represent search region. The mixed attention is defined as: 
\begin{equation}
\vspace{-2mm}
\begin{aligned}
    & k_m = {\rm Concat}(k_t, k_s), \ \ \ v_m = {\rm Concat}(v_t, v_s), \\ 
    & {\rm Attention_{t}} = {\rm Softmax}(\frac{q_{t}k_{m}^{T}}{\sqrt{d}})v_{m},\\
    & {\rm Attention}_{s} = {\rm Softmax}(\frac{q_{s}k_{m}^{T}}{\sqrt{d}})v_{m},
\end{aligned}
\end{equation}
where $d$ represents the dimension of the key, ${\rm Attention}_{t}$ and ${\rm Attention}_{s}$ are the attention maps of the target and search respectively. It contains both self attention and cross attention which unifies the feature extraction and information integration. Finally, the targets token and search token are concatenated and processed by a linear projection.

\noindent \textbf{Asymmetric mixed attention scheme.} Intuitively, the cross attention from the targets query to search area is not so important and might bring negative influence due to potential distractors. To reduce computational cost of MAM and thereby allowing for efficiently using multiple templates to deal with object deformation, we further present a customized {\em asymmetric} mixed attention scheme by pruning the unnecessary target-to-search area cross-attention. This asymmetric mixed attention is defined as follows:
\begin{small}
\begin{equation}
\vspace{-1mm}
\begin{aligned}
    & {\rm Attention_{t}} = {\rm Softmax}(\frac{q_{t}k_{t}^{T}}{\sqrt{d}})v_{t}, \\
    & {\rm Attention}_{s} = {\rm Softmax}(\frac{q_{s}k_{m}^{T}}{\sqrt{d}})v_{m}.
\end{aligned}
\end{equation}
\end{small}
In this manner, the template tokens in each MAM could remain unchanged during tracking process since it avoids influence by the dynamic search regions.
\paragraph{Discussions.}
To better expound the insight of the mixed attention, we make a comparison with the attention mechanism used by other transformer trackers.
Different with our mixed attention, TransT~\cite{tt} uses ego-context augment and cross-feature augment modules to perform self attention and cross attention progressively in two steps. Compared to the transformer encoder of STARK~\cite{stark}, our MAM shares a similar attention mechanism but with three notable differences. First, we incorporate the spatial structure information with a depth-wise convolution while they use positional encoding. More importantly, our MAM is built as a multi-stage backbone for both feature extraction and information integration, while they depend on a separate CNN backbone for feature extraction and only focus on information integration in a single stage. Finally, we also propose a different asymmetric MAM to further improve the tracking efficiency without much accuracy drop.

\subsection{MixFormer for Tracking}

\paragraph{Overall Architecture.}
Based on the MAM blocks, we build the MixFormer, a compact end-to-end tracking framework. The main idea of MixFormer is to progressively extract coupled features for target template and search area, and deeply perform the information integration between them. Basically, it comprises two components: a backbone composed of iterative target-search MAMs, and a simple localization head to produce the target bounding box.
Compared with other prevailing trackers by decoupling the steps of feature extraction and information integration, it leads to a more compact and neat tracking pipeline only with a single backbone and tracking head, without an explicit integration module or any post-processing. 
The overall architecture is depicted in Fig.~\ref{fig_arch}. 

\begin{figure}[t]
\centering
\includegraphics[width=\linewidth]{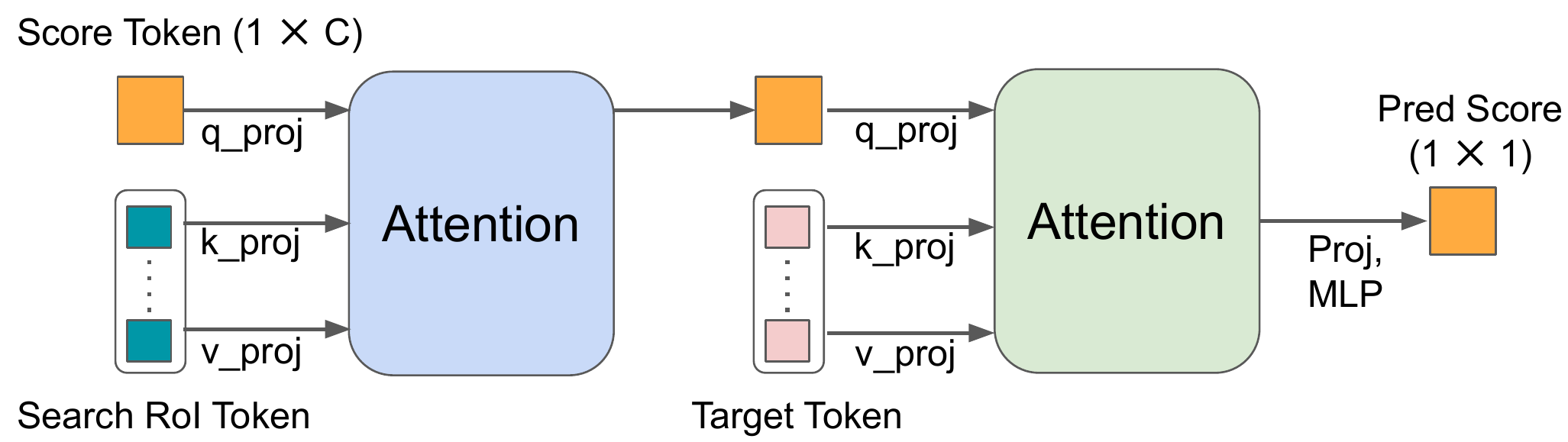}
\vspace{-6mm}
\caption{Structure of the Score Prediction Module (SPM). 
}
\vspace{-5mm}
\label{fig_score_module}
\end{figure}
\paragraph{MAM Based Backbone.}
Our goal is to couple both the generic feature extraction and target information integration within a unified transformer-based architecture. The MAM-based backbone employs a progressive multi-stage architecture design. Each stage is defined by a set of $N$ MAM and MLP layers operating on the same-scaled feature maps with the identical channel number. All stages share the similar architecture, which consists of an overlapped patch embedding layer and $N_i$ target-search mixed attention modules (i.e., a combination of MAM and MLP layers in implementation). 

Specifically, given $T$ templates (i.e., the first template and $T-1$ online templates) with the size of $T\times H_{t}\times{W_{t}}\times3$ and a search region (a cropped region according to the previous target states) with the size of $H_{s}\times{W_{s}}\times3$, we first map them into overlapped patch embeddings using a \emph{convolutional Token Embedding} layer with stride $4$ and kernel size $7$. 
The convolutional token embedding layer is introduced in each stage to grow the channel resolution while reducing the spatial resolution. 
Then we flatten the patch embeddings and concatenate them, yielding a fused token sequence with the size of $(T\times\frac{H_{t}}{4}\times\frac{W_t}{4}+\frac{H_s}{4}\times\frac{W_s}{4})\times C$, where $C$ equals to 64 or 192, $H_t$ and $W_t$ is 128, $H_s$ and $W_s$ is 320 in this work. 
After that, the concatenated tokens pass through $N_i$ target-search MAM to perform both feature extraction and target information incorporation. 
Finally, we obtain the token sequence of size $(T\times\frac{H_{t}}{16}\times\frac{W_t}{16}+\frac{H_s}{16}\times\frac{W_s}{16})\times6C$. More details about the MAM backbones could be found in the Section~\ref{model_arch} and Table~\ref{tab:arch}.
Before passed to the prediction head, the search tokens are split and reshaped to the size of $\frac{H_{s}}{16}\times\frac{W_s}{16}\times6C$. 
Particularly, we do not apply the multi-scale feature aggregation strategy, commonly used in other trackers (e.g., SiamRPN++~\cite{siamrpnPlus}, STARK~\cite{stark}).
\vspace{-3mm}
\paragraph{Corner Based Localization Head.}\label{corner_head}
Inspired by the corner detection head in STARK~\cite{stark}, we employ a fully-convolutional corner based localization head to directly estimate the bounding box of tracked object, solely with several $\rm Conv$-$\rm BN$-$\rm ReLU$ layers for the top-left and the bottom-right corners prediction respectively. At last, we can obtain the bounding box by computing the expectation over corner probability distribution~\cite{GFloss}. The difference with STARK lies in that ours is a fully convolutional head while STARK highly relies on both encoder and the decoder with more complicated design.
\vspace{-3mm}
\paragraph{Query Based Localization Head.}\label{query_head}
Inspired by DETR~\cite{detr}, we propose to employ a simple query based localization head. This sparse localization head can verify the generalization ability of our MAM backbone and yield a pure transformer-based tracking framework. Specifically, we add an extra learnable \emph{regression token} to the sequence of the final stage and use this token as an anchor to aggregate information from entire target and search area. Finally, a FFN of three fully connected layers is employed to directly regress the bounding box coordinates. This framework does not use any post-processing technique either.

\subsection{Training and Inference}
\noindent {\bf Training.}
The training process of our MixFormer generally follows the standard training recipe of current trackers~\cite{stark,transt}. We first pre-train our MAM with a CVT model~\cite{cvt}, and then fine-tune the whole tracking framework on the target dataset. 
Specifically, a combination of ${L_1}$ loss and GIoU loss~\cite{giou} is employed as follows:
\begin{equation}
    L_{loc} = \lambda_{L1} L_1(B_i, \hat{B_i}) + \lambda_{giou} L_{giou}(B_i, \hat{B_i}),
\end{equation}
where $\lambda_{L1}=5$ and $\lambda_{giou}=2$ are the weights of the two losses, $B_i$ is the ground-truth bounding box and $\hat{B_i}$ is the predicted bounding box of the targets. 

\noindent {\bf Template Online Update.}
Online templates play an important role in capturing temporal information and dealing with object deformation and appearance variations. 
However, it is well recognized that poor-quality templates may lead to inferior tracking performance. As a consequence, we introduce a score prediction module (SPM), described in Fig.~\ref{fig_score_module}, to select reliable online templates determined by the predicted confidence score. The SPM is composed of two attention blocks and a three-layer perceptron. First, a learnable \emph{score token} serves as a query to attend the search ROI tokens. It enables the score token to encode the mined target information. Next, the score token attends to all positions of the initial target token to implicitly compare the mined target with the first target. Finally, the score is produced by the MLP layer and a sigmoid activation. The online template is treated as negative when its predicted score is below than 0.5.

For the SPM training, it is performed after the backbone training and we use a standard cross-entropy loss:
\begin{equation}
\begin{split}
L_{score} = y_i{\rm log}(p_i)+(1-y_i){\rm log}(1-p_i), 
\end{split}
\end{equation}
where $y_i$ is the ground-truth label and $p_i$ is the predicted confidence score.

\noindent {\bf Inference.} During inference, multiple templates, including one static template and $N$ dynamic online templates, together with the cropped search region are fed into MixFormer to produce the target bounding box and the confidence score. We update the online templates only when the update interval is reached and select the sample with the highest confidence score. 

\section{Experiments}

\subsection{Implementation Details}
Our trackers are implemented using Python 3.6 and PyTorch 1.7.1. The MixFormer training is conducted on 8 Tesla V100 GPUs. Especially, MixFormer is a neat tracker \textbf{without} post-processing, positional embedding and multi-layer feature aggregation strategy.
\vspace{-4mm}
\begin{table*}[ht]
    \centering
    \fontsize{7pt}{3.5mm}\selectfont
    \setlength{\tabcolsep}{0.6mm}{
    \begin{tabular}{c|cccccccccc|ccc}
    \toprule
        & KCF & STM & SiamMask & D3S & SuperDiMP & AlphaRef & OceanPlus  & RPT & DualTFR & STARK & \textbf{MixFormer-1k} & \textbf{MixFormer-22k} & \textbf{MixFormer-L}\cr
        & ~\cite{kcf}& ~\cite{stm}& ~\cite{siammask} &~\cite{d3s} &~\cite{dimp} &~\cite{alpha-refine} &~\cite{oceanPlus} & ~\cite{rpt} & ~\cite{dualtfr}& ~\cite{stark} & & & \cr
        \midrule
        EAO & 0.154 & 0.308&  0.321 & 0.439 & 0.305 & 0.482 & 0.491 & 0.530 & 0.528 & 0.505 & 0.527 & \textbf{\textcolor{blue}{0.535}} & \textbf{\textcolor{red}{0.555}}\\
        Accuracy & 0.407 & 0.751 & 0.624 & 0.699 & 0.492 & 0.754 & 0.685 & 0.700 & 0.755 & 0.759& 0.746 & \textbf{\textcolor{red}{0.761}} & \textbf{\textcolor{blue}{0.762}}\\
        Robustness & 0.432 & 0.574 & 0.648 & 0.769 & 0.745 & 0.777 & 0.842 & \textbf{\textcolor{red}{0.869}} & 0.836 & 0.817 &
        0.833 & 0.854 & \textbf{\textcolor{blue}{0.855}}\\
    \bottomrule
    \end{tabular}
    }
    \vspace{-2mm}
    \caption{State-of-the-art comparison on VOT2020~\cite{vot2020}. The best two results are shown in \textbf{\textcolor{red}{red}} and \textbf{\textcolor{blue}{blue}} fonts. Our trackers use Alpha-Refine~\cite{alpha-refine} to predict masks. MixFormer-1k is pretrained with ImageNet-1k. Others are pretrained with ImageNet-22k.} 
    \label{tab:vot2020}
\vspace{-3mm}
\end{table*}

\begin{table}[pt]
\centering
\fontsize{7pt}{3.5mm}\selectfont
\setlength{\tabcolsep}{0.7mm}{
\resizebox{0.99\columnwidth}{!}{
\begin{tabular}{c|c|l|c|c}
\hline
\multicolumn{1}{l|}{}     
& Output Size
& Layer Name                                                          
& MixFormer 
& MixFormer-L \\ 
\hline

\multirow{3}{*}{Stage1}    
& $\begin{array}{c} S: 80\times80,\\ T: 32\times32
\end{array}$
& Conv. Embed.                                                     
& \multicolumn{1}{c|}{$7\times7$, $64$, stride $4$}                   
& $7\times7$, $192$, stride $4$               \\ 
\cline{2-5} 

& $\begin{array}{c} S: 80\times80,\\ T: 32\times32
\end{array}$
& \begin{tabular}[c]{@{}l@{}}MAM\\ MLP\end{tabular} 
&             
$\left[
\begin{array}{c}
     H_1 = 1 \\
     D_1=64 \\
     R_1=4
\end{array}
\right] \times 1$
&      
$\begin{bmatrix}
\begin{array}{c}
     H_1=3\\
     D_1=192\\
     R_1=4  
\end{array}
\end{bmatrix} \times 2$
\\ \hline

\multirow{3}{*}{Stage2}    
& $\begin{array}{c} S: 40\times40,\\ T: 16\times16
\end{array}$
&  Conv. Embed.
& \multicolumn{1}{c|}{$3\times3$, $192$, stride $2$}                   
& $3\times3$, $768$, stride $2$
\\ \cline{2-5} 
                           
& $\begin{array}{c} S: 40\times40,\\ T: 16\times16
\end{array}$
& \begin{tabular}[c]{@{}l@{}} MAM \\ MLP\end{tabular} 
&             
$\begin{bmatrix}
\begin{array}{c}
     H_2=3 \\
     D_2=192 \\
     R_2=4 
\end{array}
\end{bmatrix} \times 4$
&             
$\begin{bmatrix}
\begin{array}{c}
     H_2=12, \\ D_2=768,\\
     R_2=4  
\end{array}
\end{bmatrix} \times 2$
\\ \hline

\multirow{3}{*}{Stage3}    
& $ \begin{array}{c} S: 20\times20,\\ T: 8\times8
\end{array} $      
& Conv. Embed.
& \multicolumn{1}{c|}{$3\times3$, $384$, stride $2$}                   
& $3\times3$, $1024$, stride $2$
\\ \cline{2-5} 
                           
& $\begin{array}{c} S: 20\times20,\\ T: 8\times8
\end{array}$
& \begin{tabular}[c]{@{}l@{}} MAM\\ MLP\end{tabular} 
&             
$\begin{bmatrix}
\begin{array}{c}
     H_3=6 \\
     D_3=384 \\
     R_3=4  
\end{array}
\end{bmatrix} \times 16$
&
$\begin{bmatrix}
\begin{array}{c}
     H_3=16 \\
     D_3=1024 \\
     R_3=4 
\end{array}
\end{bmatrix} \times 12$
\\ \hline

\multicolumn{3}{c|}{MACs}  
& $35.61$ M
& $183.89$ M
\\ \hline
\multicolumn{3}{c|}{FLOPs}  
& $23.04$ G
& $127.81$ G
\\ \hline
\multicolumn{3}{c|}{Speed (1080Ti)}  
& $25$ FPS 
& $18$ FPS 
\\ \hline
\end{tabular}
}}
\vspace{-2mm}
\caption{MAM based backbone architectures for MixFormer and MixFormer-L. The input is a tuple of templates with shape of $128\times128\times3$ and search region with shape of $320\times320\times3$. $S$ and $T$ represent for the search region and template. $H_i$ and $D_i$ is the head number and embedding feature dimension in the $i$-th stage. $R_i$ is the feature dimension expansion ratio in the MLP layer. 
}
\label{tab:arch}
\vspace{-4mm}
\end{table}

\paragraph{Architectures.}\label{model_arch}
As shown in Table~\ref{tab:arch}, we instantiate two models, MixFormer and MixFormer-L, with different parameters and FLOPs by varying the number of MAM blocks and the hidden feature dimension in each stage. The backbone of MixFormer and MixFormer-L are initialized with the CVT-21 and CVT24-W~\cite{cvt} (first 16 layers are employed) pretrained on ImageNet~\cite{imagenet} respectively.
\vspace{-4mm}
\paragraph{Training.} 
The training set includes TrackingNet~\cite{trackingnet}, LaSOT~\cite{lasot}, GOT-10k~\cite{got10k} and COCO~\cite{coco} training dataset, which is the same as DiMP~\cite{dimp} and STARK~\cite{stark}.
While for GOT-10k test, we train our tracker by only using the GOT10k train split following its standard protocol. 
The whole training process of MixFormer consists of two stages, which contains the first 500 epochs for backbones and heads, and extra 40 epochs for score prediction head. 
We train the MixFormer by using ADAM~\cite{adam} with weight decay $10^{-4}$. The learning rate is initialized as $1e$-$4$ and decreased to $1e$-$5$ at the epoch of 400. 
The sizes of search images and templates are $320\times320$
pixels and $128\times128$ pixels respectively. For data augmentations, we use horizontal flip and brightness jittering.
\vspace{-4mm}
\paragraph{Inference.}
We use the first template and multiple online templates together with the current search region as input of MixFormer. The dynamic templates are updated when the update interval of 200 is reached by default. The template with the highest predicted score in the interval is selected to substitute the previous one.

\subsection{Comparison with the state-of-the-art trackers}
We verify the performance of our proposed MixFormer-1k, MixFormer-22k, and MixFormer-L on five benchmarks, including VOT2020~\cite{vot2020}, LaSOT~\cite{lasot}, TrackingNet~\cite{trackingnet}, GOT10k~\cite{got10k}, UAV123~\cite{uav123}.
\vspace{-4mm}
\paragraph{VOT2020.}
VOT2020~\cite{vot2020} consists of 60 videos with several challenges including fast motion, occlusion, etc.
As shown in Table~\ref{tab:vot2020}, MixFormer-L achieves the top-ranked performance on EAO criteria of 0.555, which outperforms the transformer tracker STARK with a large margin of 5\% of EAO. MixFormer-22k also outperforms other trackers including RPT (VOT2020 short-term challenge winner).

\begin{table*}[ht]
    \centering
    \fontsize{7}{9}\selectfont  
    \setlength{\tabcolsep}{2.0mm}{
    \begin{tabular}{c|ccc|ccc|ccc|cc}
    \toprule
    \multirow{2}{*}{Method} &
    \multicolumn{3}{c|}{LaSOT} &
    \multicolumn{3}{c|}{TrackingNet} &
    \multicolumn{3}{c|}{GOT-10k} &
    \multicolumn{2}{c}{UAV123} \\
    \cline{2-12}
     & AUC(\%) & $P_{Norm}$(\%) & P(\%) & AUC(\%) & $P_{Norm}(\%)$ & P(\%) & AO(\%) & $SR_{0.5}$(\%) & $SR_{0.75}$(\%) & AUC(\%) & P(\%)\\
    \midrule
    \textbf{MixFormer-L} & \textbf{\textcolor{red}{70.1}} & \textbf{\textcolor{red}{79.9}} & \textbf{\textcolor{red}{76.3}} & \textbf{\textcolor{red}{83.9}} & \textbf{\textcolor{red}{88.9}} & \textbf{\textcolor{red}{83.1}} &
    \underline{75.6} & \underline{85.7} & \underline{72.8} &
    69.5 & \textbf{\textcolor{blue}{91.0}} \\
    \textbf{MixFormer-22k} & \textbf{\textcolor{blue}{69.2}} & \textbf{\textcolor{blue}{78.7}} & \textbf{\textcolor{blue}{74.7}} & \textbf{\textcolor{blue}{83.1}} & \textbf{\textcolor{blue}{88.1}} & \textbf{\textcolor{blue}{81.6}}
    & \underline{72.6} & \underline{82.2} & \underline{68.8} & \textbf{\textcolor{red}{70.4}} & \textbf{\textcolor{red}{91.8}} \\
    \textbf{MixFormer-1k} & 67.9 & 77.3 & 73.9 & 82.6 & 87.7 & 81.2 & \underline{73.2} & \underline{83.2} & \underline{70.2} & 68.7 & 89.5 \\ 
     \textbf{MixFormer-22k*} & - & - & - & - & - & - & \textbf{\textcolor{blue}{70.7}} & \textbf{\textcolor{red}{80.0}} & \textbf{\textcolor{red}{67.8}} & - & - \\ 
    \textbf{MixFormer-1k*} & - & - & - & - & - & - & \textbf{\textcolor{red}{71.2}} & \textbf{\textcolor{blue}{79.9}} & \textbf{\textcolor{blue}{65.8}} & - & - \\ 
    \hline
    STARK~\cite{stark} & 67.1 & 77.0 & - & 82.0 & 86.9 & - & 68.8 & 78.1 & 64.1 & - & - \\
    KeepTrack~\cite{keeptrack} & 67.1 & 77.2 & 70.2 & - & - & - & - & - & - & \textbf{\textcolor{blue}{69.7}} & - \\
    DTT~\cite{dtt} & 60.1 & - & - & 79.6 & 85.0 & 78.9 & 63.4 & 74.9 & 51.4 & - & - \\
    SAOT~\cite{saot} & 61.6 & 70.8 & - & - & - & - & 64.0 & 75.9 & - & - & - \\
    AutoMatch~\cite{automatch} & 58.2 & - & 59.9 & 76.0 & - & 72.6 & 65.2 & 76.6 & 54.3 & - & - \\
    TREG~\cite{treg} & 64.0 & 74.1 & - & 78.5 & 83.8 & 75.0 & 66.8 & 77.8 & 57.2 & 66.9 & 88.4 \\
    DualTFR~\cite{dualtfr} & 63.5 & 72.0 & 66.5 & 80.1 & 84.9 & - & - & - & - & 68.2 & - \\
    TransT~\cite{transt} & 64.9 & 73.8 & 69.0 & 81.4 & 86.7 & 80.3 & 67.1 & 76.8 & 60.9 & 69.1 & - \\
    TrDiMP~\cite{tmt} & 63.9 & - & 61.4 & 78.4 & 83.3 & 73.1 & 67.1 & 77.7 & 58.3 & 67.5 & - \\
    STMTracker~\cite{stmtrack} & 60.6 & 69.3 & 63.3 & 80.3 & 85.1 & 76.7 & 64.2 & 73.7 & 57.5 & 64.7 & - \\
    SiamR-CNN~\cite{siamrcnn} & 64.8 & 72.2 & - & 81.2 & 85.4 & 80.0 & 64.9 & 72.8 & 59.7 & 64.9 & 83.4 \\
    PrDiMP~\cite{prdimp} & 59.8 & 68.8 & 60.8 & 75.8 & 81.6 & 70.4 & 63.4 & 73.8 & 54.3 & 68.0 & - \\
    OCEAN~\cite{ocean} & 56.0 & 65.1 & 56.6 & - & - & - & 61.1& 72.1 & 47.3 & - & - \\
    FCOT~\cite{fcot} & 57.2 & 67.8 & - & 75.4 & 82.9 & 72.6 & 63.4 & 76.6 & 52.1 & 65.6 & 87.3 \\
    SiamGAT~\cite{siamgat} & 53.9 & 63.3 & 53.0 & - & - & - & 62.7 & 74.3 & 48.8 & 64.6 & 84.3 \\
    CGACD~\cite{CGACD} & 51.8 & 62.6 & - & 71.1 & 80.0 & 69.3 & - & - & - & 63.3 & 83.3 \\
    SiamFC++~\cite{siamfc++} & 54.4& 62.3 & 54.7 & 75.4 & 80.0 & 70.5 & 59.5& 69.5 & 47.9 & - & - \\
    MAML~\cite{maml} & 52.3 & - & - & 75.7 & 82.2 & 72.5 & - & - & - & - & - \\
    D3S~\cite{d3s} & - & - & - & 72.8 & 76.8 & 66.4 & 59.7 & 67.6 & 46.2 & - & - \\
    DiMP~\cite{dimp} & 56.9 & 65.0 & 56.7 & 74.0 & 80.1 & 68.7 & 61.1 & 71.7 & 49.2 & 65.4 & - \\
    ATOM~\cite{atom} & 51.5 & 57.6 & 50.5 & 70.3 & 77.1 & 64.8 & 55.6 & 63.4 & 40.2 & 64.3 & - \\
    SiamRPN++~\cite{siamrpnPlus} & 49.6 & 56.9 & 49.1 & 73.3 & 80.0 & 69.4 & 51.7 & 61.6 & 32.5 & 61.0 & 80.3 \\
    MDNet~\cite{mdnet} & 39.7 & 46.0 & 37.3 & 60.6 & 70.5 & 56.5 & 29.9 & 30.3 & 9.9 & 52.8 & - \\
    SiamFC~\cite{siamfc} & 33.6 & 42.0 & 33.9 & 57.1 & 66.3 & 53.3 & 34.8 & 35.3 & 9.8 & 48.5 & 69.3 \\
    \bottomrule
    \end{tabular}}
    \vspace{-2mm}
    \caption{State-of-the-art comparison on TrackingNet~\cite{trackingnet}, LaSOT~\cite{lasot}, GOT-10k~\cite{got10k} and UAV123~\cite{uav123}. The best two results are shown in \textbf{\textcolor{red}{red}} and \textbf{\textcolor{blue}{blue}} fonts. The \underline{underline} results of GOT-10k are not considered in the comparison, since the models are trained with datasets other than GOT-10k. MixFormer-1k is the model pretrained with ImageNet-1k. Others are pretrained with ImageNet-22k. * denotes for trackers trained only with GOT-10k train split.}
    \vspace{-5mm}
    \label{tab:resultsl}
\end{table*}
\vspace{-4mm}
\paragraph{LaSOT.}
LaSOT~\cite{lasot} has 280 videos in its test set. We evaluate our MixFormer on the test set to validate its long-term capability. 
The Table~\ref{tab:resultsl} shows that our MixFormer surpasses all other trackers with a large margin. Specifically, MixFormer-L achieves the top-ranked performance on NP of 79.9\%, surpassing STARK by 2.9\% even without multi-layers feature aggregation. 
\vspace{-4mm}
\paragraph{TrackingNet.}
TrackingNet~\cite{trackingnet} provides over 30K videos with more than 14 million dense bounding box annotations. The videos are sampled from YouTube, covering target categories and scenes in real life. We validate MixFormer on its test set. From Table~\ref{tab:resultsl}, we find that our MixFormer-22k and MixFormer-L set a new state-of-the-art performance on the large scale benchmark.
\vspace{-4mm}
\paragraph{GOT10k.}
GOT10k~\cite{got10k} is a large-scale dataset with over 10000 video segments and has 180 segments for the test set. Apart from generic classes of moving objects and motion patterns, the object classes in the train and test set are zero-overlapped. As shown in Table~\ref{tab:resultsl}, our MixFormer-GOT obtain state-of-the-art performance on the test split.
\paragraph{UAV123.}
UAV123~\cite{uav123} is a large dataset containing 123 Sequences with average sequence length of 915 frames, which is captured from low-altitude UAVs. Table~\ref{tab:resultsl} shows our results on UAV123 dataset. Our MixFormer-22k and MixFormer-L outperforms all other trackers.

\subsection{Exploration Studies}
To verify the effectiveness and give a thorough analysis on our proposed MixFormer, we perform a detailed ablation study on the large-scale LaSOT dataset.
\vspace{-4mm}
\paragraph{Simultaneous process vs. Separate process.} 
As the core part of our MixFormer is to unify the procedure of feature extraction and target information integration, we compare it to the separate processing architecture (e.g. TransT~\cite{tt}). The comparison results are shown in Table~\ref{tab_component} \#1, \#2, \#3 and \#8. Experiments of \#1 and \#2 are end-to-end trackers comprising a self-attention based backbone, ${n}$ cross attention modules to perform information integration and a corner head. \#3 is the tracker with CVT as backbone and TransT's ECA+CFA(4) as interaction. Experiment of \#8 is our MixFormer {\bf without} multiple online templates and asymmetric mechanism, denoted by \emph{MixFormer-Base}. MixFormer-Base largely increases the model of \#1 (using one CAM) and \#2 (using three CAMs) by 8.6\% and 7.9\% with smaller parameters and FLOPs. This demonstrates the effectiveness of unified feature extraction and information integration, as both of them would benefit each other.
\vspace{-3mm}
\begin{table}[pt]
    \centering
    \fontsize{7.5}{8.5}\selectfont  
    \setlength{\tabcolsep}{0.8mm}{
    \vspace{-1mm}
    \begin{tabular}{c|ccc|ccc}
        \hline
        \text{\#}&Backbone&Integration&Head&Params.&FLOPs&AUC\\
        \hline
        1& SAM(21) & CAM(1) & Corner & 37.35M & 20.69G & 59.8 \\
        2& SAM(21) & CAM(3) & Corner & 40.92M & 22.20G & 60.5 \\
        3& SAM(21) & ECA+CFA(4) & Corner & 49.75M & 27.81G & 66.9 \\
        \hline
        \hline
        4& SAM(20)+MAM(1) & - & Corner & 35.57M & 19.97G & 65.8 \\
        5& SAM(15)+MAM(6) & - & Corner & 35.67M & 20.02G & 66.2 \\
        6& SAM(10)+MAM(11) & - & Corner & 35.77M & 20.07G & 67.4 \\
        7& SAM(5)+MAM(16) & - & Corner & 35.87M & 20.12G & 68.1 \\
        \hline
        \hline
        8& MAM(21) & - & Corner & 35.97M & 20.85G & \cellcolor{gray!20}\textbf{68.4} \\
        9& MAM(21) & - & Query & 31.46M & 19.13G & 66.0 \\
        \hline
    \end{tabular}
    }
    \vspace{-1mm}
    \caption{Analysis of the MAM based framework. '-' denotes the component is not used. SAM represents for self attention module, CAM for cross attention module and MAM for the proposed mixed attention module. The numbers in brackets represent the number of the blocks. Performance is evaluated on LaSOT.}
    \label{tab_component}
\vspace{-6mm}
\end{table}

\vspace{-2mm}
\paragraph{Study on stages of MAM.}
To further verify the effectiveness of the MAMs, we conduct experiments as in Table~\ref{tab_component} \#4, \#5, \#6, \#7 and \#8, to investigate the performance of different numbers of MAM in our MixFormer. We compare our MAM with the self-attention operations (SAM) with out cross-branch information communication. We find that more MAMs contribute to higher AUC score.
It indicates that extensive target-aware feature extraction and hierarchical information integration play a critical role to construct an effective tracker, which is realized by the iterative MAM. Especially, when the number of MAM reaches 16, the performance reaches 68.1, which is comparable to the MixFormer-Base containing 21 MAMs.
\begin{table}[pt]
    \centering
    \setlength{\tabcolsep}{1mm}{
    \vspace{-1mm}
    \small
    \begin{tabular}{c|c|cc}
        \hline
        &Asymmetric.&FPS (1080Ti)&AUC\\
        \hline
        MixFormer-Base& - & 19 & \cellcolor{gray!20}\textbf{68.4} \\
        MixFormer-Base& \checkmark & 25 & 68.1 \\
        \hline
    \end{tabular}
    }
    \vspace{-2mm}
    \caption{Ablation for asymmetric mixed attention mechanism.}
\vspace{-1mm}
\label{tab_asym}
\end{table}

\begin{table}[pt]
\small
    \centering
    \setlength{\tabcolsep}{1mm}{
    \vspace{-1mm}
    \begin{tabular}{c|cc|c}
        \hline
        &Online & Score&AUC\\
        \hline
        MixFormer& - & - & 68.1 \\
        MixFormer& \checkmark & - & 66.6 \\
        MixFormer& \checkmark & \checkmark & \cellcolor{gray!20}\textbf{69.2} \\
        \hline
    \end{tabular}
    }
    \vspace{-2mm}
    \caption{Ablation for online templates update mechanism.}
    \label{tab_online}
\vspace{-1mm}
\end{table}

\begin{table}[pt]
\small
    \centering
    \fontsize{6pt}{2.5mm}\selectfont
    \setlength{\tabcolsep}{1mm}{
    \vspace{-1mm}
    \small
    \begin{tabular}{c|cc|c}
        \hline
        & Pretrain & Train &AUC\\
        \hline
        MixFormer& ImageNet-1k & Whole &  67.9 \\
        MixFormer& ImageNet-22k& Whole & \cellcolor{gray!20}\textbf{69.2} \\
        MixFormer& ImageNet-22k& GOT-10k & 62.1 \\
        \hline
    \end{tabular}
    }
    \vspace{-2mm}
    \caption{Study on pretraining and training datasets. 'Whole' denotes for using the whole datasets including GOT-10k, LaSOT, TrackingNet and COCO.}
    \label{tab_data}
\vspace{-7mm}
\end{table}

\vspace{-4mm}
\paragraph{Study on localization head.}
To verify the generalization ability of our MAM backbone, we evaluate the MixFormer-Base with two types of localization head as described in Section~\ref{corner_head} (fully convolutional head vs. query based head). The results are shown as in Table~\ref{tab_component} \#8 and \#9 for the corner head and the query-base head respectively.
MixFormer-Base with the fully convolutional corner head outperforms that of the query-based head. Especially, MixFormer-Base with corner head surpass all the other state-of-the-art trackers even without any post-processing and online templates. 
Besides, MixFormer-Base with the query head, a pure transformer-based tracking framework, obtains a comparable AUC of 66.0 with STARK-ST and KeepTrack~\cite{keeptrack} and far exceed query-head STARK-ST of 63.7.
It demonstrates the generalization ability of our MAM backbone.
\vspace{-4mm}
\paragraph{Study on asymmetric MAM.}
The asymmetric MAM is used to reduce computational cost and allows for usage of multiple templates during online tracking. As shown in Table~\ref{tab_asym}, the asymmetric MixFormer-Base increases the running speed of 24\% while achieving a comparable performance, which demonstrates asymmetric MAM is important for building an efficient tracker.
\vspace{-4mm}
\paragraph{Study on online template update.}
As demonstrated in Table~\ref{tab_online}, MixFormer with online templates, sampled by a fixed update interval, performs worse than that with only the first template, and the online MixFormer with our score prediction module achieves the best AUC score. It suggests that selecting reliable templates with our score prediction module is of vital importance.
\vspace{-4mm}
\paragraph{Study on training and pre-training datasets.}
To verify the generalization ablility of our MixFormer, we conduct an analysis on different pre-training and training datasets, as shown in Table~\ref{tab_data}. MixFormer pretrained by ImageNet-1k still outperforms all the SOTA trackers (e.g., TransT~\cite{tt}, KeepTrack~\cite{keeptrack}, STARK~\cite{stark}), even without post-processing and multi-layer feature aggregation. In addition, MixFormer trained with GOT-10k also achieves an impressive AUC of 62.1, which outperforms a majority of trackers trained with the whole tracking datasets.

\begin{figure}[pt]
\centering
\includegraphics[width=\linewidth]{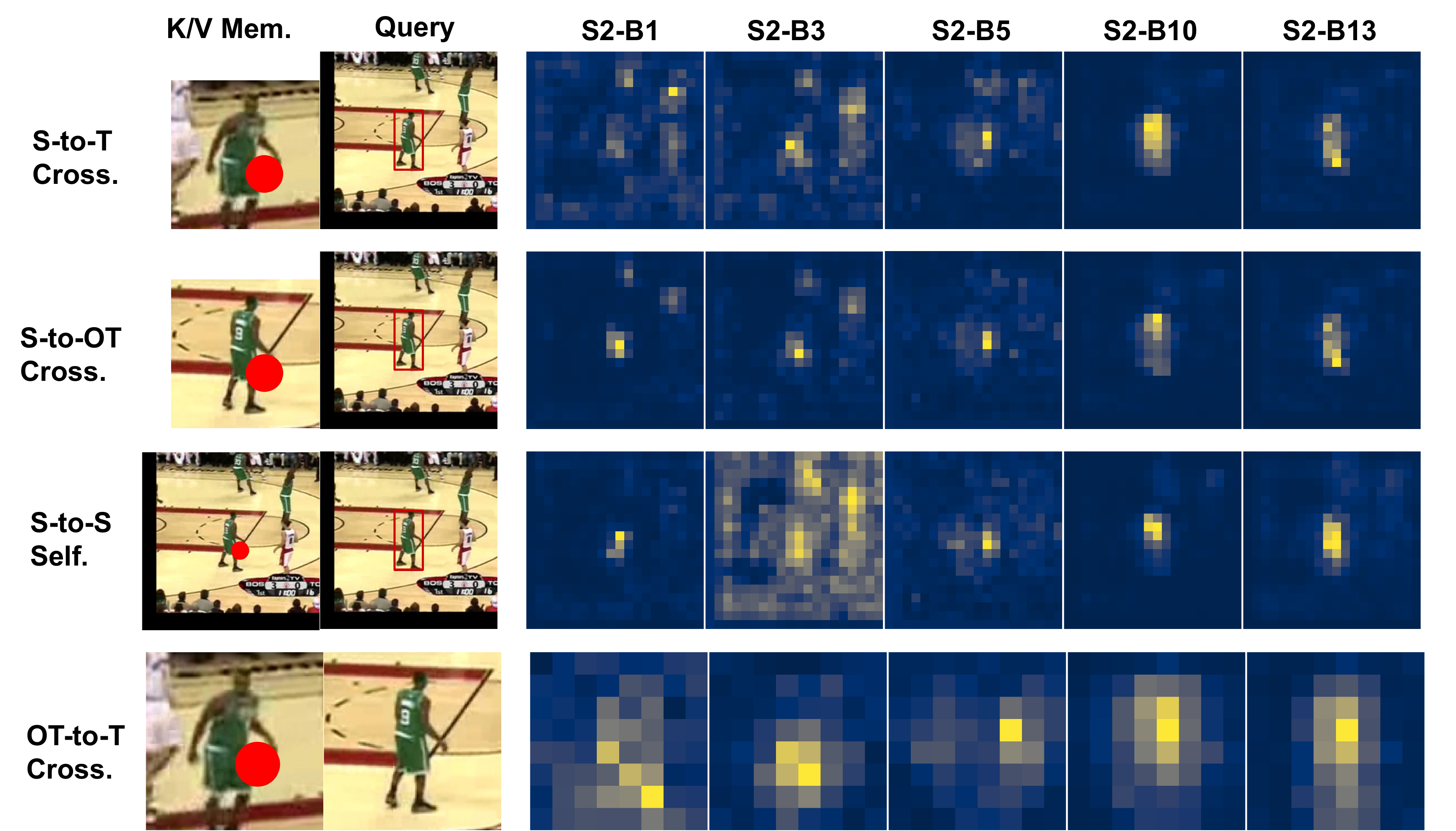}
\vspace{-4mm}
\caption{Visualization results of different attention weights on \textit{Basketball} of OTB100. \textbf{S-to-t} is search-to-template cross attention, \textbf{S-to-OT} is search-to-online-template cross attention, \textbf{S-to-S} is self attention of search region and \textbf{OT-to-T} is online-template-to-template cross attention. \textbf{S$i$-B$j$} represents for Stage-$i$ and Block-$j$ of MixFormer. Best viewed with zooming in.}
\vspace{-6mm}
\label{fig:vis_attn_bas}
\end{figure}

\vspace{-4mm}
\paragraph{Visualization of attention maps.}
\label{vis_attn}
To explore how the mixed attention works in MixFormer backbone, we visualize some attention maps in Fig.~\ref{fig:vis_attn_bas}. From the four types of attention maps, we derive that: (\romannumeral1) distractors in background get suppressed layer by layer, (\romannumeral2) online templates may be more adaptive to appearance variation and help to discriminate the target, (\romannumeral3) the foreground of multiple templates can be augmented by mutual cross attention, (\romannumeral4) a certain position tends to interact with the surrounding local patch.

\section{Conclusion}
We have presented MixFormer, an end-to-end tracking framework with iterative mixed attention, aiming to unify the feature extraction and target integration and result in a neat and compact tracking pipeline. Mixed attention module performs both feature extraction and mutual interaction for target template and search area. In empirical evaluation, MixFormer shows a notable improvement over other prevailing trackers for short-term tracking. In the future, we consider extending MixFormer to multiple object tracking.

\vspace{1mm}
\small \noindent {\bf Acknowledgement.} This work is supported by National Natural Science Foundation of China  (No.62076119, No.61921006),  Program for Innovative Talents and Entrepreneur in Jiangsu Province, and Collaborative Innovation Center of Novel Software Technology and Industrialization.

\section*{Appendix}

\begin{figure*}[pt]
\centering
\includegraphics[width=0.9\linewidth]{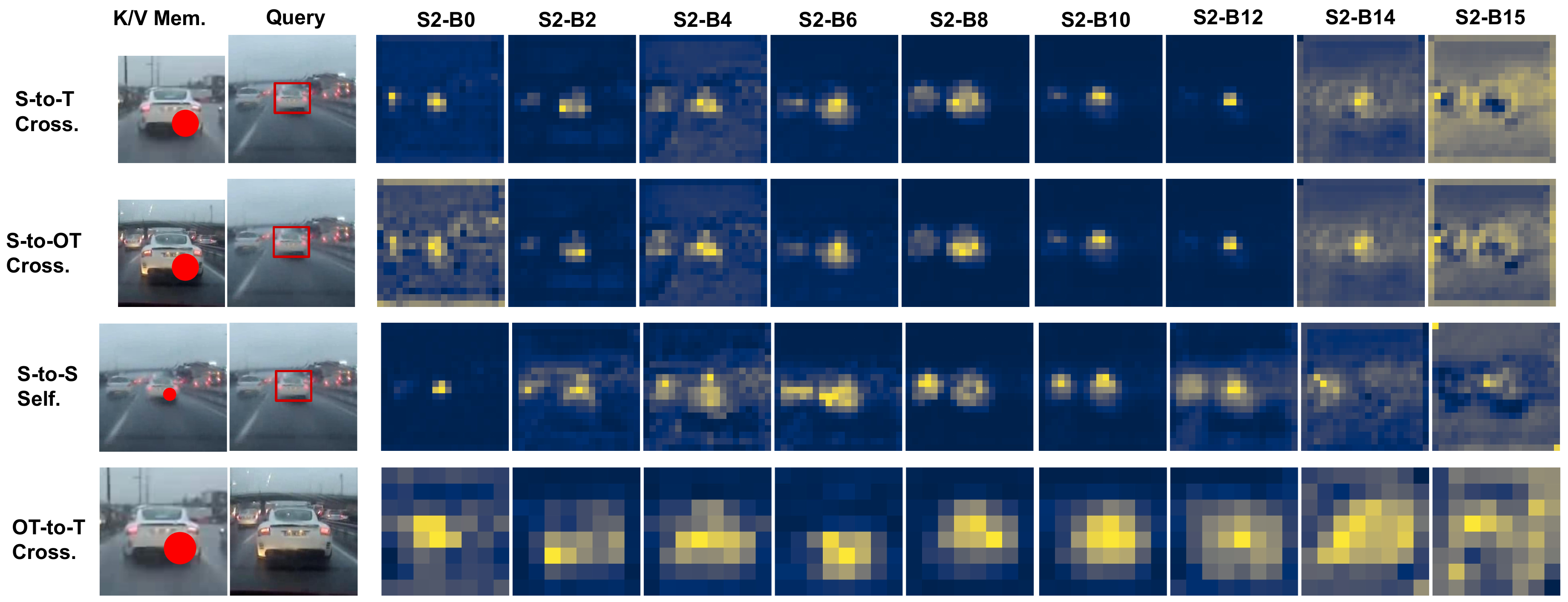}
\vspace{-3mm}
\caption{Visualization results of different attention weights on \textit{car-2} of LaSOT. \textbf{S-to-t} is search-to-template cross attention, \textbf{S-to-OT} is search-to-online-template cross attention, \textbf{S-to-S} is self attention of search region and \textbf{OT-to-T} is online-template-to-template cross attention. \textbf{S$i$-B$j$} represents for Stage-$i$ and Block-$j$ of MixFormer. Best viewed with zooming in.}
\vspace{-5mm}
\label{fig:vis_attn_car-2}
\end{figure*}

In this appendix, we first provide more results and analysis on OTB100~\cite{otb} and LaSOT~\cite{lasot} datasets. Then we give more visualization results of the attention weights on LaSOT. Finally, we provide more training details.

\section*{A. More Results}

\begin{figure}[pb]
\centering
\includegraphics[width=\linewidth]{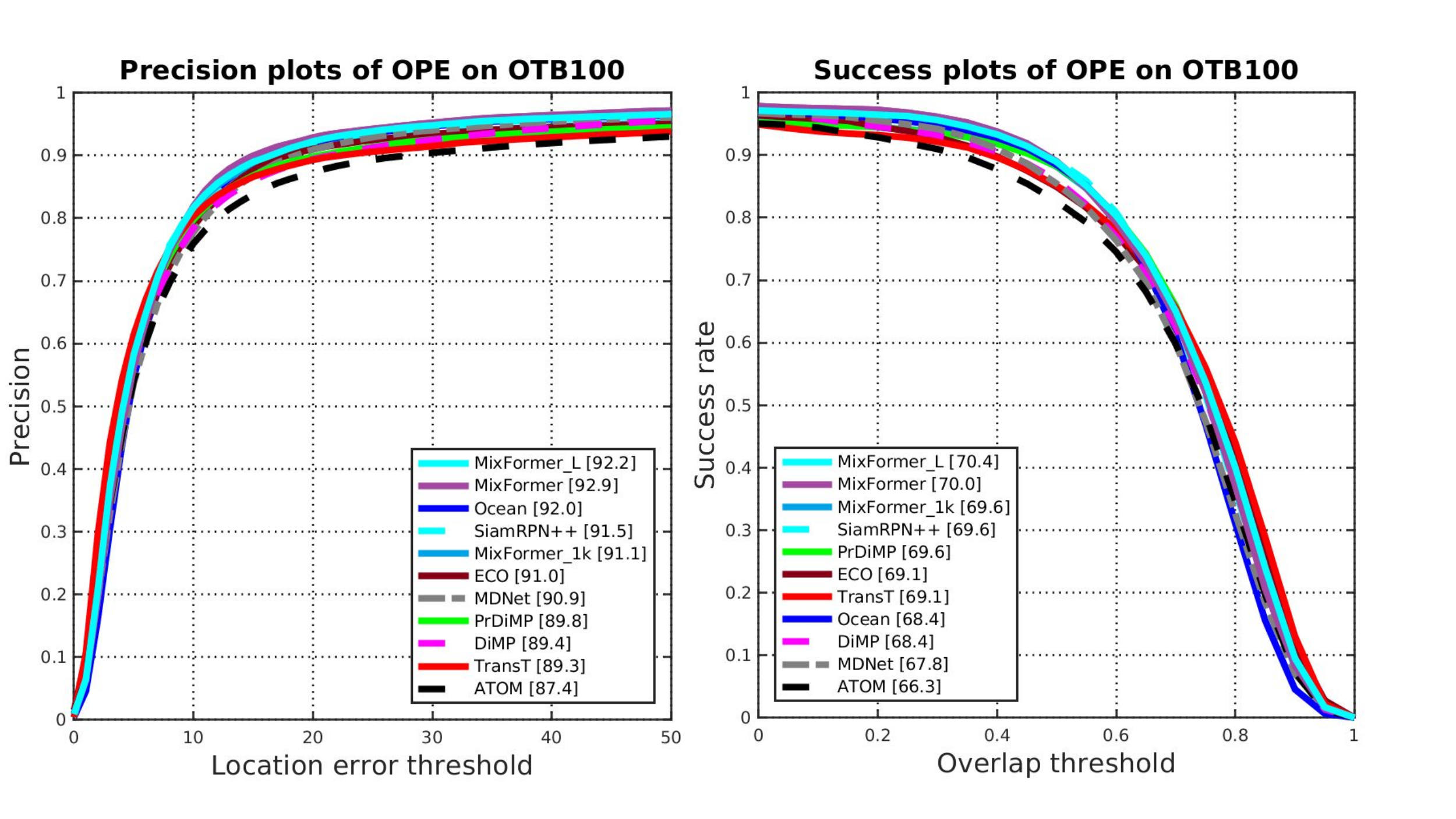}
\caption{State-of-the-art comparison on the OTB100 dataset. Best viewed with zooming in.}
\label{fig:otb}
\end{figure}

\paragraph{OTB-100.}
OTB100~\cite{otb} is a commonly used benchmark, which evaluates performance on Precision and AUC scores. Figure.~\ref{fig:otb} presents results of our trackers on both two metrics on OTB-100 benchmark. MixFormer-L reaches competitive performance w.r.t. state-of-the-art trackers, surpassing the transformer tracker TransT by 1.3\% on AUC score. Besides, MixFormer-L is slightly higher than MixFormer.

\begin{figure}[pt]
\centering
\includegraphics[width=\linewidth]{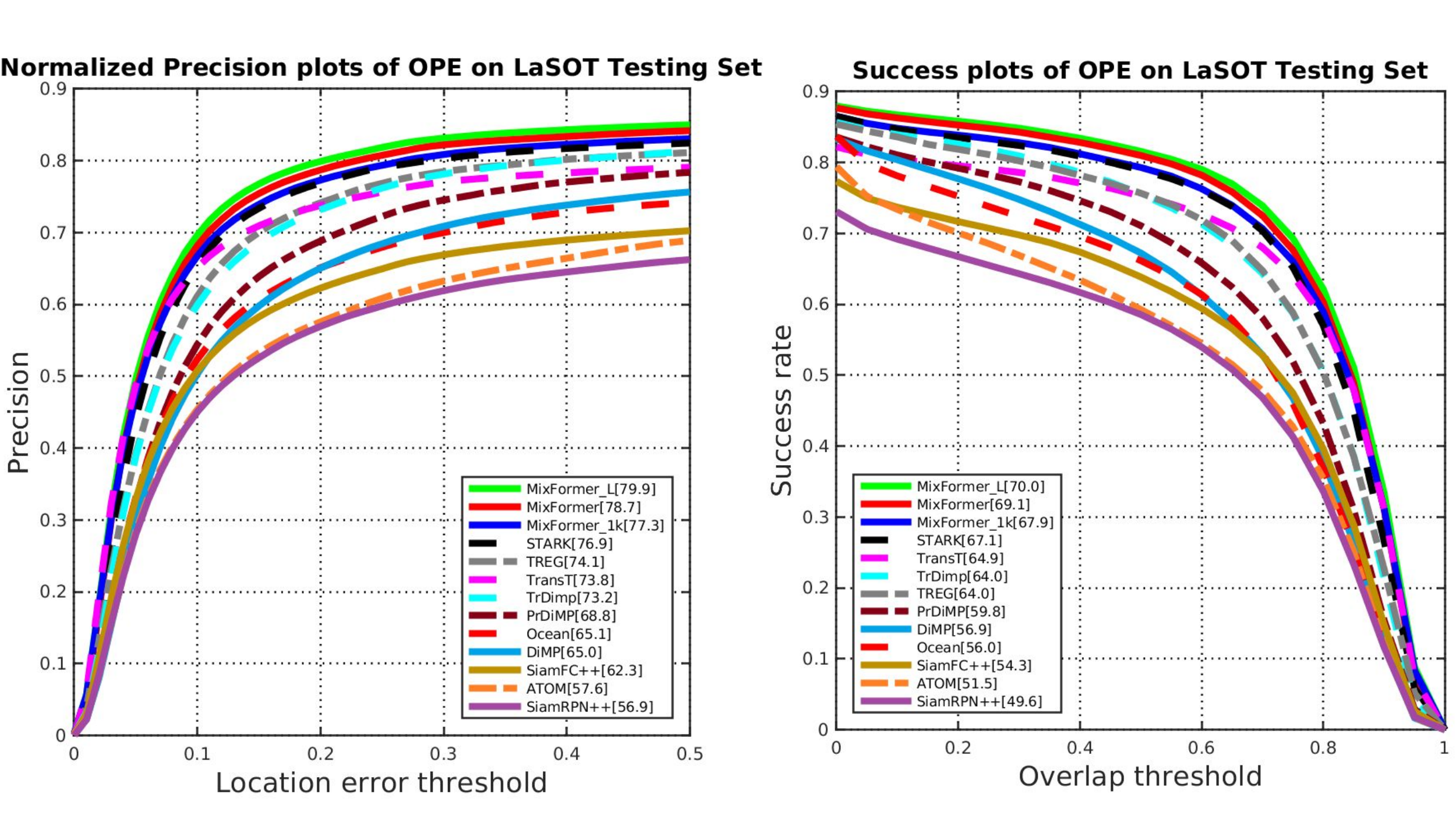}
\vspace{-8mm}
\caption{State-of-the-art comparison on the LaSOT dataset.}
\vspace{-5mm}
\label{fig:lasot}
\end{figure}

\paragraph{LaSOT.}
LaSOT~\cite{lasot} has 280 videos in its test set. We evaluate our MixFormer on the test set to validate its long-term capability. To give a further analysis, we provide Success plot and Precision plot for LaSOT in Fig.~\ref{fig:lasot}. It proves that improvement is due to both higher accuracy and robustness. 

\section*{B. More Visualization Results}
In this section, we provide more visualization results of attention weights on \textit{car-2} of LaSOT test dataset in Fig.~\ref{fig:vis_attn_car-2}. From the example, we can arrive at the same conclusion with section~\ref{vis_attn}. Besides, from the last two lines, we infer that the features of last two blocks tend to adapt to the bounding box prediction head.

\section*{C. Training Details}

We propose a 320x320 search region plus two 128x128 input images to make a fair comparison with prevailing trackers (e.g., Siamese-based trackers, STARK and TransT). Generally, we use 8 Tesla V100 GPUSs to train MixFormer with batch size of 32. MixFormer can also be trained on 8 2080Ti GPUs having only 11GB memory, with batch size of 8 per GPU. We use CvT21 and CvT24-W as the pretrained model for MixFormer and MixFormer-L respectively. We apply gradient clip strategy with the clip normalization rate of 0.1. For training stage-1 of MixFormer (i.e., MixFormer without SPM), we use GIoU loss and $L_1$ loss, with the weights of 2.0 and 5.0 respectively. Besides, the Batch Normalization layers of MixFormer backbone are frozen during the whole training process. For SPM training process, the backbone and corner-based localization head are frozen and the batch size is 32. SPM is trained for 40 epochs with the learning rate decays at 30 epochs.

{\small
\bibliographystyle{ieee_fullname}
\bibliography{egbib}
}

\end{document}